\documentclass[conference]{IEEEtran}
\IEEEoverridecommandlockouts
\usepackage{cite}
\usepackage{amsmath,amssymb,amsfonts}
\usepackage{algorithmic}
\usepackage{graphicx}
\usepackage{textcomp}
\usepackage{xcolor}
\def\BibTeX{{\rm B\kern-.05em{\sc i\kern-.025em b}\kern-.08em
    T\kern-.1667em\lower.7ex\hbox{E}\kern-.125emX}}

\usepackage[utf8]{inputenc}
\usepackage{times}
\usepackage{soul}
\usepackage{url}
\usepackage[hidelinks]{hyperref}
\usepackage[utf8]{inputenc}
\usepackage[small]{caption}
\usepackage{booktabs}
\usepackage{algorithm}

\usepackage{amsthm}

\newtheorem{theorem}{Theorem}

\usepackage{nomencl}
\makenomenclature

\newcount\Comments  
\usepackage{color}
\definecolor{darkgreen}{rgb}{0,0.5,0}
\definecolor{purple}{rgb}{1,0,1}
\newcommand{\kibitz}[2]{\ifnum\Comments=1\textcolor{#1}{#2}\fi}
\newcommand{\kizito}[1]{\kibitz{red}      {[Kizito: #1]}}

\newcommand{\xingyu}[1]  {\kibitz{darkgreen}   {[Xingyu: #1]}}
\newcommand{\lorenzo}[1]  {\kibitz{purple}   {[Lorenzo: #1]}}

\begin{document}


\title{Assessing the Safety and Reliability of Autonomous Vehicles from Road Testing\thanks{ This work is partially supported by the UK EPSRC through the Offshore Robotics for Certification of Assets (ORCA) Hub [EP/R026173/1].}}

\author{\IEEEauthorblockN{Xingyu Zhao, Valentin Robu, David Flynn}
\IEEEauthorblockA{\textit{School of Engineering \& Physical Sciences} \\
\textit{Heriot-Watt University}\\
Edinburgh, U.K. \\
\{xingyu.zhao,v.robu,d.flynn\}@hw.ac.uk}
\and
\IEEEauthorblockN{Kizito Salako, Lorenzo Strigini}
\IEEEauthorblockA{\textit{Centre for Software Reliability} \\
\textit{City, University of London}\\
London, U.K. \\
\{k.o.salako,l.strigini\}@city.ac.uk}
}


\maketitle

\begin{abstract}
There is an urgent societal need to assess whether autonomous vehicles (AVs) are safe enough. From published quantitative safety and reliability assessments of AVs, we know that, given the goal of predicting very low rates of accidents, road testing alone requires infeasible numbers of miles to be driven. However, previous analyses do not consider any knowledge prior to road testing – knowledge which could bring substantial advantages if the AV design allows strong expectations of safety before road testing. We present the advantages of a new variant of Conservative Bayesian Inference (CBI), which uses prior knowledge while avoiding optimistic biases. 
We then study the trend of disengagements (take-overs by human drivers) by applying Software Reliability Growth Models (SRGMs) to data from Waymo's public road testing over 51 months, in view of the practice of software updates during this testing. Our approach is to not trust any specific SRGM, but to assess forecast accuracy and then improve forecasts. We show that, coupled with accuracy assessment and recalibration techniques, SRGMs could be a valuable test planning aid.

\end{abstract}


\section{Introduction}

In recent years, autonomous vehicles (AVs) have moved rapidly from labs to public roads. AVs are claimed to have the potential to make road traffic much safer and more efficient. Much research has been conducted on various aspects of deploying AVs, e.g. design, implementation, regulation and legal issues \cite{anderson_autonomous_2016,paden_survey_2016,fagnant_preparing_2015,koopman_autonomous_2017,bonnefon_social_2016,schwarting_planning_2018}. Due to considerable investment, practical AVs seem just around the corner; e.g., Waymo LLC -- formerly the Google self-driving car project -- launched its first commercial AV taxi service on 5th December 2018.

Prior to that, Waymo, like other AV manufacturers, has been testing its AVs on public roads in the U.S. for years. Such operational testing in real traffic, with close observation of AV performance, is a necessary part of assessing the safety of AVs. Indeed, Google presented its 1.4 million miles of road testing data as important testimonial evidence in the U.S. Congress hearings on AV regulation \cite{urmson_hands_2016}. Meanwhile, scholars \cite{banerjee_hands_2018,kalra_driving_2016}  have used the same kind of data to draw sobering conclusions about how far AVs are from achieving their safety goals and (an even harder challenge) demonstrating that it is achieved.


These studies mostly rely on descriptive statistics, giving insights on various aspects of AV safety \cite{banerjee_hands_2018,favaro_autonomous_2018,dixit_autonomous_2016,lv_analysis_2018}. A RAND Corporation study \cite{kalra_driving_2016} has been highly cited, and in this paper we refer to it for comparison, to illustrate similarities and differences between alternative statistical approaches to assessment and the results thereof. For the reader's convenience, we will refer to this paper as ``the RAND study''. The RAND study uses classical statistical inference to find how many miles need to be driven to claim a desired AV reliability with a certain confidence level. However, such techniques do not address how safety and reliability claims\footnote{In this paper we only deal with probabilistic claims, so ``reliability'' claims will be about probabilities of occurrence of failures, ``safety'' claims about failures that are safety-relevant. The two kinds do not require different statistical reasoning, except as far as affected by practical  differences in e.g. frequencies, desired bounds, and  degrees of observability.}, based on operational testing evidence, can be made in a way that:

\textit{a) is practical given very rare failure events}, such as fatalities and crashes. If and when AVs achieve their likely safety targets, rates of such events will be very small, say a $10^{-7}$  \textit{probability of a crash event per mile (pcm)}. Especially for test failures that could cause a fatality -- counted to estimate future \textit{probability of a fatality event per mile} (\textit{pfm}) -- most companies will observe no such failures, if they are even close to the target. If they did observe any, the required redesign/update of the AV could make the fatality data obsolete. Gaining confidence in such low 
failure rates is a major challenge \cite{littlewood_validation_1993,butler_infeasibility_1993}, possibly requiring infeasible amounts of operation to discriminate between the conjectures that the (say) \textit{pfm}  is as low as desired, or is not. This was the case in the RAND study findings.

\textit{b) incorporates relevant prior knowledge}. In conventional system, this prior knowledge would typically include evidence of soundness of design (as supported by verification results) and quality of process. AVs rely for core functionalities on machine learning (ML) systems, for which the ability to prove correct design is lacking (despite intense research). But AVs, just as more conventional systems, will normally include safety precautions (e.g. defence in depth design with safety monitors/watchdogs \cite{littlewood_reasoning_2012}). Indeed, such ``safety subsystems'' are not only suggested in policy making \cite{anderson_autonomous_2016}, but also extensively implemented by AV manufacturers \cite{waymo_waymo_2018,amnon_shashua_plan_2017}. Such safety subsystems have relatively simple functionalities (e.g. bringing the vehicle to a safe stop), can avoid relying on ML functions, and allow for conventional verification methods.
If these safety subsystems are the basis for prior confidence in safety, evidence about their development and verification should be combined (in a statistically principled way) with operational testing evidence. The same applies if evidence for the ML functions or the whole system is available (e.g. from automated testing \cite{tian_deeptest_2018} or formal verification \cite{kamali_formal_2017,fisher_verifiable_2018}).

\textit{c) considers that 
while road testing data are collected, the AVs are being updated}. For an unchanging vehicle operating under statistically unchanging conditions, ``constant event rate'' models, as applied, e.g., in the RAND study, may apply. However, there is an expectation that an AV's ML-based core systems improve as the vehicle evolves with testing experience, which should be reflected in the frequency of failure-related events.
So, for instance, one would expect a decreasing trend in the frequency of \emph{disengagements}\footnote{Failures causing AVs' control to be switched to human drivers.}, as has been observed. E.g., \cite{banerjee_hands_2018} reports noticeable changes for \textit{disengagements per mile} (\textit{dpm}) over cumulative miles. 
Although decreasing \textit{dpm} does not imply increasing reliability/safety of the AV\footnote{Interpreting \textit{dpm} as an indicator of AV safety is wrong \cite{banerjee_hands_2018} and potentially dangerous, through both being misleading and creating incentives to improve  \textit{dpm} rather than safety \cite{koopman_safety_2019}. Proper use of \textit{dpm} data in arguing safety would require assessing the interplay between (a) the evolution of ML functions, (b) that of the safety drivers, and (c) the safety subsystems.

Note that, an improvement of ML-based functions most likely reduces drivers' ability to trigger disengagements when needed, by affecting e.g. their trust in the AV and situation awareness. Also, the probability of a safety subsystem's successful action depends on the probability distribution of the demands created by the ML-based functions \cite{PopovStrigini2010ISSRE}.}\lorenzo{LS to put in something more concrete: DONE but dropped the next few sentences for now... data about fatalities/crashes are relatively sparse, and we have no idea of the proportion being mitigated by human supervisors who will not be there for fully-autonomous cars. So how the truth of AV safety and reliability evolves or remains over time is still an open question. We believe it depends on both the reliability measure under study and the particular design of an AV (e.g. the different safety architectures adopted by manufactures), which requires more rigorous scientific studies in future.}, it is a useful indicator to study (e.g. for planning of road testing, and as inputs to more refined analysis of actual AV reliability).
Assessment of changing measures like \textit{dpm} should use statistical approaches that account for such changes. 


The key contributions of this work are:

\textit{a) For constant safety and reliability scenarios}, we develop a new \emph{Conservative Bayesian Inference} (CBI) framework for reliability assessment, that can incorporate both failure-free and rare failures evidence. Including the case of non-zero failure counts generalises existing CBI methods \cite{bishop_toward_2011,strigini_software_2013,zhao_modeling_2017,zhao_conservative_2015}, applied in other settings such as nuclear safety, that consider only failure-free evidence. For AVs, instead, occasional failures are to be expected.
So our new framework incorporates failures into the assessment. Being a Bayesian approach, it also allows for the incorporation of prior knowledge of non-road-testing evidence (e.g. verified aspects of the behaviour of an AV’s ML algorithms; verification results for the safety subsystems). We then compare claims based on our CBI framework with claims from other AV case studies, using the same data and settings (in particular, we consider how CBI compares with the well-known inference approach used in the RAND study). CBI shows how these other approaches can be either optimistic, or too pessimistic, and the difference may be substantial.


\textit{b) For scenarios with the AV evolving over cumulative miles driven},
we show how past AV disengagement data can be used to predict future disengagement, and such predictions evaluated against observations. To this end, we use \emph{Software Reliability Growth Models (SRGMs)} \cite{Miller1986EOS}. 
Fitting these models to Waymo's publicly available testing data, we evaluate the accuracy of their reliability forecasts, and show how the models' predictions can be improved by ``recalibration'' -- a model improvement technique that utilizes statistical data on how the models' past predictions fall short of observed outcomes \cite{brocklehurst_techniques_1996}.
 
The outline of the rest of this paper is as follows. Next, we present (Section \ref{sec_OT_and_failure_process}) preliminaries on assessing reliability from operational testing. Section \ref{sec_CBI} details the new CBI framework while Section \ref{sec_SRGM} introduces SRGMs, applied to Waymo's disengagement data. Sections \ref{sec_related_work} and \ref{sec_conclusions} summarise related work, contributions and future work.

\section{Operational Testing \& Failure Processes}
\label{sec_OT_and_failure_process}
For conventional safety-critical systems, statistical evaluation from operational testing, or ``proven in use'' arguments, are part of standards like IEC61508 \cite{iec_61508_2010} and EN50129\cite{en50129_railway_2003}.
These practices are supported by established \cite{atwood2003handbook,strigini_guidelines_1997}\lorenzo{perhaps add the NASA document} and still evolving  \cite{walter_bayesian_2017,bishop_deriving_2017,utkin_imprecise_2018} probabilistic methods. Since, for AVs, road testing is emphasised as evidence for proving safety and reliability, it is not surprising that inference methods using such operational evidence are attracting attention.

In general, depending on the system under study, a stochastic failure process is chosen as a mathematical abstraction of reality. Here, for AVs, we describe the failure processes (of fatalities, crashes or disengagements) as:

\textit{a) Bernoulli processes} for the occurrence of fatalities or crashes. 
These models assume the probability of a failure\footnote{For brevity, we call ``failure'' generically the event of interest (disengagement, crash, etc.), and use ``failure rate'' both in its technical meaning as the parameter (\textit{dpm}) of, say, a Poisson process, and for the probability of failure per mile in a Bernoulli model (\textit{pfm}, \textit{pcm}).} per driven mile is a constant, and events in one mile are independent of events in any other mile driven. This process assumption may not really hold for various reasons (e.g. AV reliability can evolve during testing, or AVs required to operate under dependent, changing road/environmental conditions). For some of these objections, it can be observed that in many practical scenarios a Bernoulli model is an acceptable approximation of the more complex, real process.\kizito{\bf TODO: include Bernoulli process use-case references} Even for such a scenario, one would still expect that changing the ML-based systems during testing would make the Bernoulli model inapplicable. Arguments for still using it as a first approximation could be, for instance, that the non-ML based safety subsystems raise the overall AV reliability to a much higher level than that of the ML-based systems, and this overall AV reliability remains constant during observation, despite the evolution of the ML-based systems\footnote{``A first approximation'' because the evolution of the ML-based core changes the set of failures to be tolerated by the safety subsystem (\emph{cf} \cite{PopovStrigini2010ISSRE}). A previous statistical study \cite{favaro_examining_2017} found that some key AV reliability measures, e.g. \textit{pcm} for AVs, appear constant over time but this is not enough to support making it a modelling \emph{assumption}.}.\\ 
There are two reasons for us to use this model: i) the model is simple enough to highlight the challenges of AV safety assessment, and ii) for the purpose of comparison against the RAND study \cite{kalra_driving_2016} which uses this model.

\textit{b) Point processes} for disengagements: Point processes, such as Poisson processes (in which inter-event times are independent, identically distributed, exponential random variables) are well-suited for modelling reliability during continuous system operation. Another example, that of Non-homogeneous Poisson processes, allows for non-stationarity and dependence in the failure data \cite{cinlarStocProcBook}. In what follows, using families of point processes from the SRGM literature, we illustrate how the predictive accuracy of forecasts of future AV disengagements can be evaluated, and possibly improved (see Sec.~\ref{sec_SRGM}).


\section{The CBI Approach for \textit{pfm} \& \textit{pcm} claims}
\label{sec_CBI}

Published CBI methods \cite{bishop_toward_2011,strigini_software_2013,zhao_conservative_2015,zhao_modeling_2017,zhao_conservative_2018} are for conventional safety-critical software (e.g. nuclear protection systems where any failure is assumed to have significant consequences), and thus deal with operational testing where \textit{no} failures occur. However, AI systems do fail in operation. For AVs, although very rare, crashes and a fatality have been reported. To deal with (infrequent) failures, we propose a more general CBI framework, in which $0$ failures becomes a special case. 
For AVs, we apply CBI to assessing \textit{pfm} and \textit{pcm}, and compare the results with those of the RAND study. 


Assessment claims using statistical inference come in different flavours. The RAND study derives ``classical'' confidence statements about the claim of an acceptable failure rate. E.g., $95\%$ confidence in a bound of $10^{-x}$ means that if the failure rate were greater than $10^{-x}$, the chances of observing no failures in the miles driven would be $5\%$ at most. 
The Bayesian approach, instead, treats failure rate as a random variable with a ``prior'' probability distribution (``prior'' to test observations). The prior is updated (via Bayes' theorem) using test results, giving a ``posterior'' distribution. Decisions are based on probabilities derived from the posterior distribution, e.g. the probability (``Bayesian confidence''), say 0.95, of the failure rate being less than $10^{-x}$. These two notions of confidence have radically different meanings, but
decision making based on levels of ``confidence'' of either kind is common: hence we will compare the amounts and kinds of evidence required to achieve high ``confidence'' with either approach.

Now, a challenge for using Bayesian inference in practice is the need for complete prior distributions (of the failure rate, in the present problem).
A common way to deal with this issue is to choose distribution functions that seem plausible in the domain and/or mathematically convenient (e.g. for conjugacy). 
However, often, such a distribution does not describe only one's prior knowledge, but adds extra, unjustified assumptions. This may do no harm if the posterior depends on the data much more than on the prior distribution, but in our case (with few or zero failures), the conclusions of the inference will be seriously sensitive to these assumptions: those extra assumptions risk dangerously unsound reasoning. 


CBI bypasses this problem: rather than a \textit{complete} prior distribution, an assessor is more likely to have (and be able to justify)\lorenzo{
}
more limited \textit{partial prior knowledge}, e.g. a prior confidence bound -- ``I am 80\% confident that the failure rate is smaller than $10^{-3}$'' -- based on e.g. experience with results of similar quality practices in similar projects.
This partial prior knowledge is far from a complete prior distribution. Rather, it \textit{constrains} the prior: there is an \textit{infinite set} of prior distributions satisfying the constraints. Then, CBI determines the \textit{most conservative} one from this set, in the sense of minimising the posterior confidence on a reliability bound.


\subsection{CBI With Failures in Testing}

As described in Section \ref{sec_OT_and_failure_process}, consider a Bernoulli process representing a succession of miles driven by an AV, and let $X$ be the unknown \emph{pfm} value (the setup if, instead, crashes are considered, is analogous). Suppose $k$ failures in $n$ driven miles are observed (denoted as $k\&n$ for short in equations). If $F(x)$ is a prior distribution function for $X$ then, for some stated reliability bound $p$,
\begin{equation}
\label{eq_post_cf_bound_with_complete_prior}
Pr(X \leqslant p\mid k\&n)=\frac{\int_{0}^{p} x^k(1-x)^{n-k} \mathrm dF(x) }{\int_{0}^{1} x^k(1-x)^{n-k} \mathrm dF(x)}
\end{equation}

As an example, suppose that, rather than some complete prior distribution, only partial prior beliefs are expressed about an AV's \emph{pfm}:
\begin{equation}
\label{eq_prior_constraints_1}
Pr(X \leqslant \epsilon)=\theta,\quad Pr(X\geqslant p_l)=1
\end{equation} 
The interpretations of the model parameters are:

$\bullet \enspace \epsilon$ is the engineering goal, a target safety level that developers try to satisfy for a given reliability measure (e.g. \textit{pfm}). To illustrate, for \textit{pfm}, this goal could be two orders \cite{liu_how_2019}, or three orders \cite{amnon_shashua_plan_2017}, of magnitude safer than human drivers.


$\bullet \enspace \theta$ is the prior confidence that the engineering goal has been achieved \emph{before} testing the AVs on public roads. Such prior confidence could be obtained from simulations, or from verification of the AV safety subsystems, and has to be high enough to decide to proceed with public road testing.

$\bullet \enspace p_l$ is a lower bound on the failure rate: the best reliability claim feasible given current vehicle technology. 
For instance, \textit{pfm} cannot be smaller than, say $10^{-15}$, due to catastrophic hardware failures (e.g. tyre/engine fails on a highway), \emph{even if the AV's ML-based systems are perfect}. Research assuming inevitable fatalities, e.g. \cite{awad_moral_2018}, supports such $p_l$.

The foregoing is just one interpretation of the parameters; interpretations can vary between manufacturers and across business models. 

Now, assuming one has the prior beliefs \eqref{eq_prior_constraints_1}, the following CBI theorem shows what these beliefs allow one to rigorously claim about an AV's safety and reliability.
\begin{theorem} A prior distribution that minimises \eqref{eq_post_cf_bound_with_complete_prior} subject to the constraints \eqref{eq_prior_constraints_1} is the two-point distribution, $Pr(X=x)=\theta{\bf 1}_{x=x_1} + (1-\theta){\bf 1}_{x=x_3}\,$, where $p_l\leqslant x_1 \leqslant \epsilon < x_3\,$, and the values of $x_1$ and $x_3$ both depend on the model parameters (i.e. $p_l, \epsilon, p$) as well as $k$ and $n$. Using this prior, the smallest value for \eqref{eq_post_cf_bound_with_complete_prior} is
	\allowdisplaybreaks\begin{align}
	\frac{x_1^k(1-x_1)^{n-k}\theta}{x_1^k(1-x_1)^{n-k}\theta + x_3^k(1-x_3)^{n-k}(1-\theta)}{\bf 1}_{p>\epsilon}
	\label{eq_CBI_post_conf_bound_see_failures}
	\end{align}
	where ${\bf 1}_{\tt S}$ is an indicator function -- it is equal to 1 when {\tt S} is true and 0 otherwise.
	\label{thrm_1}
\end{theorem}
The proof of Theorem 1 is in appendix \ref{sec_app_A}. Depicted in Fig.~\ref{fig_two_point_priors} are two common situations (given different values of the model parameters): with failure-free and rare failures evidence.

\begin{figure}[htbp!]
	\centering
	\includegraphics[width=1\linewidth]{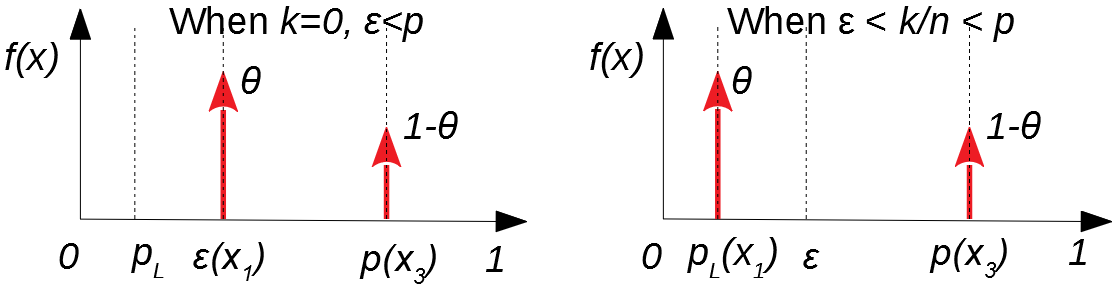}
	\caption{Conservative two-point priors for two choices of model parameters -- with failure free data (left) and rare failures (right).}
	\label{fig_two_point_priors}
\end{figure}

Solving \eqref{eq_CBI_post_conf_bound_see_failures} for $n$ -- the miles to be driven to claim the \emph{pfm} is less than $p$ with probability $c$, upon seeing $k$ failures -- provides our main technical result. From a Bayesian perspective, $n$ will depend on the prior knowledge \eqref{eq_prior_constraints_1}. In what follows, we compare the proposed $n$ values from CBI, the RAND study, a Uniform prior and Jeffreys prior (as suggested by regulatory guidance like \cite{atwood2003handbook}). Similar comparisons can be made for \textit{pcm}; we omit these due to page limitations.

\subsection{Numerical Examples of CBI for \textit{pfm} Claims}
\label{sec_num_examples_CBI}

In the RAND study, data from the U.S. department of transportation supported a \textit{pfm} for human drivers of $1.09e{-8}$ in 2013. For illustration, suppose that a company aims to build AVs two orders of magnitude safer, i.e. $\epsilon=1.09e{-10}$, as proposed by \cite{liu_how_2019}. Also, assume $p_l=10^{-15}$: that is, the unknown \textit{pfm} value cannot be better than $10^{-15}$.

\textbf{Q1: How many fatality-free miles need to be driven to claim a \textit{pfm} bound at some confidence level?}

\begin{figure}[htbp!]
	\centering
	\includegraphics[width=1\linewidth]{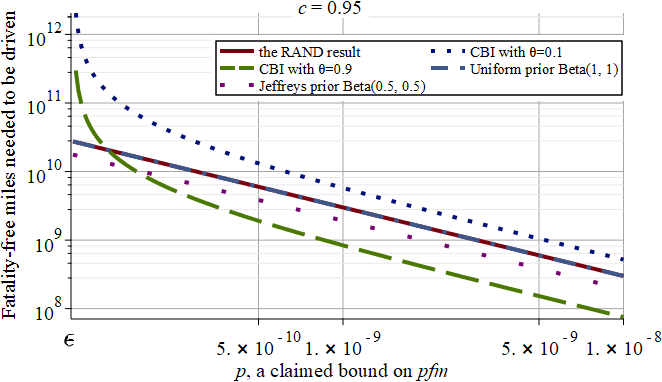}
	\caption{Fatality-free miles needed to be driven to demonstrate a \textit{pfm} claim with $95\%$ confidence. Note, the curves for Bayes with a uniform prior and the RAND results overlap in the figure (to be exact, there is a constant difference of 1 between them which is simply a consequence of the similarity between their analytical expressions in this scenario).}
	\label{fig_pfm_no_failure}
\end{figure}

With the prior knowledge \eqref{eq_prior_constraints_1}, we answer Q1 by setting $k=0$ and solving \eqref{eq_CBI_post_conf_bound_see_failures} for $n$. Fig.~\ref{fig_pfm_no_failure} shows the CBI results with $\theta\!=\!0.1$ (weak belief) and $\theta\!=\!0.9$ (strong belief) respectively, compared with the RAND results, and Bayesian results with a uniform prior {\tt Beta}$(1,1)$ and the Jeffreys prior for Binomial models ({\tt Beta}$(0.5,0.5)$ \cite[p.6.37]{atwood2003handbook}).
Fig.~\ref{fig_pfm_no_failure} shows that \eqref{eq_CBI_post_conf_bound_see_failures} can imply significantly more, or less, miles must be driven than suggested by either the RAND study or the other Bayesian priors -- depending on how confident one is \emph{before seeing test results} that the goal $\epsilon$ has been reached. For instance, to claim, with 95\% confidence, that AVs are as safe as human drivers (so $p=1.09e{-8}$), the RAND analysis requires 275 million fatality-free miles, whilst CBI with $\theta=0.9$ only requires 69 million fatality-free miles, with 90\% prior confidence that the AVs are two orders of magnitude safer than humans (based on, e.g., having the core ML-based systems backed up by non-ML safety channels that are relatively simple and easier to be verified. Such verification can be the case in traditional safety-critical systems \cite{littlewood_reasoning_2012}).\kizito{``what'' is the case for traditional safety-critical systems? The possibility of verification of such non-ML safety channels or that safety critical systems pair up ML and non-ML, but verifiable, subsystems?} 

Alternatively, if one has only a ``weak'' prior belief in the engineering goal being met ($\theta=0.1$), then CBI requires 476 million fatality-free miles -- significantly more than the other approaches compared.

The reader should not be surprised that our conservative approach does not always prescribe more fatality-free miles be driven than that prescribed by the RAND study -- different decision criteria and statistical inference methods can yield different results from the same data \cite{berger_could_2003}. However, it is true that, for any confidence $c$, CBI will require significantly more miles than the RAND study prescriptions for all claims $p$ ``close enough'' to the engineering goal $\epsilon$.

\begin{figure}[htbp!]
	\centering
	\includegraphics[width=1\linewidth]{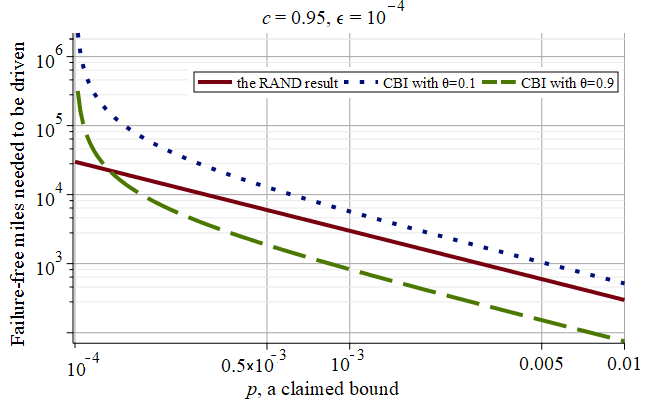}
	\caption{Failure-free miles needed to be driven to demonstrate a less stringent reliability claim with $95\%$ confidence.}
	\label{fig_less_stringent_claim}
\end{figure}

We note that, for AVs that may have less stringent reliability requirements (e.g. AVs doing regular inspection missions on offshore rigs), both the engineering goal and reliability claims can be much less stringent than the examples in Fig.~\ref{fig_pfm_no_failure}. We present CBI and RAND results for such a scenario in Fig.~\ref{fig_less_stringent_claim}, with an engineering goal $\epsilon=10^{-4}$ and a range $[10^{-4} , 10^{-2}]$ for the claimed bound $p$. Although it shows the same pattern as Fig.~\ref{fig_pfm_no_failure}, the evidence required to demonstrate a reliability claim being met with the given confidence level is much less and within a feasible range. For instance, when the claim of interest is $p=10^{-3}$, CBI with a strong prior belief in the engineering goal being met (i.e. $\theta=0.9$) requires less than $10^3$ failure-free miles, while the RAND method requires 2 to 3 times as many.

Notice that, for all of the scenarios we have presented so far, no amount of testing will support trust in any bound $p$ lower than $\epsilon$. This is because of constraint \eqref{eq_prior_constraints_1}. It allows a range of possible prior distributions -- and thus posterior confidence bounds -- but with no added basis for trusting any bound better than $\epsilon$ (as exemplified in Fig.~\ref{fig_pfm_no_failure}). Hence, a conservative decision maker that has partial prior knowledge \eqref{eq_prior_constraints_1} 
cannot accept a claim, on the basis of the fatality-free operation, that the AV reliability exceeds the engineering goal.
Of course, if \emph{further} evidence justifies a prior knowledge in some bound $p$ ($<\epsilon$), then CBI can give more informative claims.

\textbf{Q2: How many miles need to be driven, with fatality events, to claim a \emph{pfm} bound at some confidence level?}

The RAND study answers this question via classical hypothesis testing, choosing as an example a confidence bound 20\% better than human drivers' \textit{pfm} in 2013.
Their result (in number of miles required) is shown in boldface in Table.~\ref{tab_miles_to_drive_with_failures}.

In the Bayesian approach, posterior confidence depends on observations: in order to compare with the RAND study result, we thus postulate an observed number of fatalities consistent with the RAND study analysis. As an example, we consider that, given a \textit{pfm} equal to the above confidence bound, and driving the number of miles found necessary in the RAND study, the expected number of fatalities would be
$k=8.72e{-9} \times 4.97e9 \approx 43$ (where $8.72e{-9}$ is a reliability claim obtained from $4.97e9$ fatality free miles in the RAND model).
We thus assume 43 fatalities and show in column 1 of Table~\ref{tab_miles_to_drive_with_failures} the miles required by the Bayesian approaches, including CBI, Uniform and Jeffreys priors. In addition to the purpose of comparison, this case also represents a long term scenario in which, as popularity and public use of AVs grow, the count of fatal accidents progressively reaches high values. 
We show what evidence would then be needed to reassure the public that reliability claims are still being met.
\lorenzo{all to re-read this above bit with the 'progressively"}

For a short term scenario, as a second example, the last column of Table~\ref{tab_miles_to_drive_with_failures} shows the corresponding results, if only one fatality occurs. Again, we compare the results of classical hypothesis testing, CBI and using other Bayesian priors.



All of the examples in Table~\ref{tab_miles_to_drive_with_failures} ``agree'': the miles needed to make these claims are prohibitively high. However, given the CBI prior beliefs, the CBI numbers \emph{require 10$\sim$20 times more miles than the rest if 43 fatalities are seen}. The number at the bottom of column 1 represents the miles needed to demonstrate that, after fatalities consistent with \emph{pfm}$=8.72 e{-9}$, there is only a $5\%$ chance of the true \textit{pfm} being worse than that.
The difference from the RAND results may seem large, but it is in the interest of public safety: CBI avoids implicit, unwittingly optimistic assumptions in the prior distribution.

We recall that with no fatalities, the CBI example \textit{does} offer a sound basis for achieving high confidence with substantially fewer test miles than the RAND approach requires (e.g. 69 \textit{vs} 275 million miles).

\lorenzo{"prohibitively'' does the job... for the non-safety version let's insert a "real wold description": "most commentators consider this prohibitevly high, e.g. $10^9$ miles means running 10,000 cars for 100,000 miles each...  you can drive them at 50mph average in 2000 hours, or less than 3 months." Prohibitive, sure.. 2500 cars on the road for one year.} 



\begin{table}[htbp!]
	\centering
	\resizebox{0.9\columnwidth}{!}{%
		\begin{tabular}{l|c|c}  
			\toprule
			&\textit{p}=8.72e-9, \textit{k}=43 & \textit{p}=4.12e-9, \textit{k}=1\\
			\midrule
			Classical & $\mbox{\bf 4.97e9}$  & $2.43e8$      \\
			Uniform priors & $6.40e9$  &   $1.15e{9}$\\
			Jeffreys priors & $6.33e9$  &  $9.48e8$ \\
			CBI with $\theta=0.9$  & $7.89e10$  & $3.88e9 $     \\
			\bottomrule
		\end{tabular}
	}
	\caption{Miles needed to support a \emph{pfm} claim $p$ with 95\% confidence, with $k$ fatalities.}
	\label{tab_miles_to_drive_with_failures}
\end{table}

\textbf{Q3: How many more fatality-free miles need to be driven to compensate for one newly observed fatality?}

This question relates to a plausible scenario in the case of accidents\footnote{A recent example is the Uber AV crash in Arizona.}:
an AV has been driven for $n_1$ fatality-free miles, justifying a \textit{pfm} claim, say $p$ (with a fixed confidence $c$), via CBI based on this evidence and some given prior knowledge. Then suddenly a fatality event happens. Instead of redesigning the system (as no evidence exists to point to a technical/AI control design fault), the company still believes in its prior knowledge, attributes the fatality to ``bad luck'' and asks to be allowed more testing to prove its point.
If  the public/regulators accept this request,
it is useful to know how many extra fatality-free miles, say $n_2$, are needed to compensate for the fatality event, so that the company can demonstrate the same reliability $p$ with confidence $c$.

\begin{figure}[bhtp!]
	\centering
	\includegraphics[width=1\linewidth]{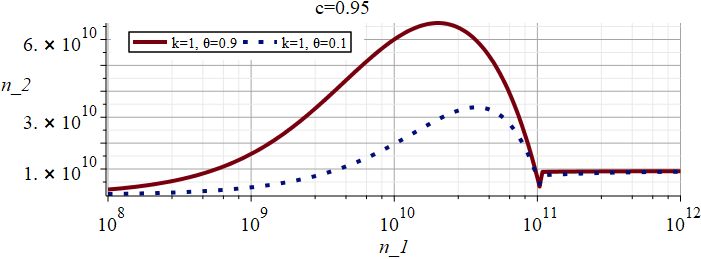}
	\caption{Fatality-free miles needed to compensate one newly observed fatality given $n_1$ fatality-free miles has been driven before.}
	\label{fig_extra_miles_needed}
\end{figure}

To answer this, apply the CBI model in two steps (fixing the confidence level $c$ and prior knowledge $\theta$): (i) determine the claim $[ X\!\leqslant\!p\,]$ that $n_1$ will support with $k\!=\!0$ (i.e. fix $k,n$ \& solve \eqref{eq_CBI_post_conf_bound_see_failures} for $p$). (ii) determine the miles that support the claim $[X\!\leqslant\!p\,]$ upon seeing $k\!=\!1$ (i.e. fix $k,p$ \& solve \eqref{eq_CBI_post_conf_bound_see_failures} for $n$). Then $n_2 \!=\! n - n_1$ more fatality-free miles are needed to compensate for the fatality; we plot some scenarios in Fig.~\ref{fig_extra_miles_needed}.

The solid curve in Fig.~\ref{fig_extra_miles_needed} shows a uni-modal pattern, decreasing as $n_1$ approaches the value $n^\ast=1.06e{11}$ (with a corresponding $p$ value, $p^\ast=1.16e\!-\!10$, derived from the 1st step), then increasing again with an asymptote of $n_2=1/\epsilon$, as $n_1$ goes to infinity. A complete formal analysis deriving $p^\ast$ and the asymptote of $n_2=1/\epsilon$ is in Appendix \ref{sec_app_B}.

Intuitively, the more fatality-free miles were driven, the higher one's confidence in reliability; and thus, the more miles needed to restore that confidence after a fatality occurs. But, if $n_1$ was such as to allow confidence in a claim close to $p_1$, then after the fatality, a much smaller $n_2$ is needed to be able to claim $p_1$ again. As $n_1$ tends to infinity, interestingly, there is a ceiling on the required $n_2$, \textit{for all values} of $c$ and $\theta$. We note that the shape of the curve (including the asymptote on the right) is invariant with respect to $c$ and $\theta$.
\kizito{NOT for safety version: give an intuitive explanation of shape. } 
%
%
%
%

\section{SRGMs for \textit{dpm} Predictions}
\label{sec_SRGM}
Whilst the previous sections focused on very rare events like fatalities and crashes, in this section we focus on a metric for a more frequent event, often reported for AV road testing data: \emph{disengagements per mile} (\textit{dpm}). Several descriptive statistical studies for \textit{dpm} exist: e.g., Banerjee and co-authors, using large-scale AV road testing data, show negative correlation between \textit{dpm} and cumulative miles driven over three years, but still not reaching AV manufacturers' targets despite millions of miles driven  \cite{banerjee_hands_2018}.
As part of road-test planning, any forecast of future \textit{dpm} must account for this trend of apparent improvement.

The idea behind Software Reliability Growth Models (SRGMs) is that each fault contributes to causing failures stochastically during operation. When a failure occurs, the software is updated in an attempt to fix that fault, then use of the software, or testing, resumes until the next failure reveals another fault. During this fault-finding and fixing process, recorded inter-failure times are used to calibrate probabilistic models so as to extrapolate the trend, in probabilistic terms, e.g. predicting the mean, or median, time to the next failure.

Many SRGMs have been developed, based on different assumptions (e.g. how much each fault contributes to the overall failure rate). Comparing them by how plausible their assumptions seem has not proven good guidance, and no single SRGM proved universally accurate \cite{abdel-ghaly_evaluation_1986}. As an alternative, techniques were proposed \cite{littlewood_new_1992} to assess and compare SRGMs' prediction accuracy \textit{over the history of a specific product}. One could thus choose which SRGM to trust, or even ``recalibrate'' them to improve predictions for that system. Thus, the best practice is to apply multiple SRGMs to the failure data of the system under study, recalibrating them as appropriate, and compare the prediction accuracy, so that we can gradually learn which SRGM seems to be best for the current prediction needs \cite{littlewood_validation_1993}.



Statistical properties of AVs, such as \textit{dpm}s, exhibit growth, as training/self-learning is applied after failures occur.
We apply various SRGMs to \textit{disengagement} data, and assess the models' predictive accuracy. The latter seems even more necessary for AVs, with their ML-based systems, than for conventional software-based systems,
as knowledge of AVs' learning mechanisms is so imperfect (and often not available to third-party assessors) that we cannot choose \textit{a priori} the most fit SRGM for a given AV.


\subsection{Applying SRGMs to Waymo AVs Data}

The California AV Testing Regulations require annual reports on disengagements from every manufacturer authorised to test AVs on public roads. We applied SRGMs to the data reported from Waymo covering 51 months of testing,
available from Waymo\footnote{www.dmv.ca.gov/portal/dmv/detail/vr/autonomous/testing} at the time of writing. 
We use PETERS, a state-of-the-art toolset that implements 8 SRGMs,
recalibration, comparison and visualisation techniques. We select the most trustworthy SRGM to predict, after each failure (i.e. disengagement), the median miles to next disengagement (MMTD), based on the series of previous inter-failure mile data.

\begin{figure}[htbp!]
	\centering
	\includegraphics[width=1\linewidth]{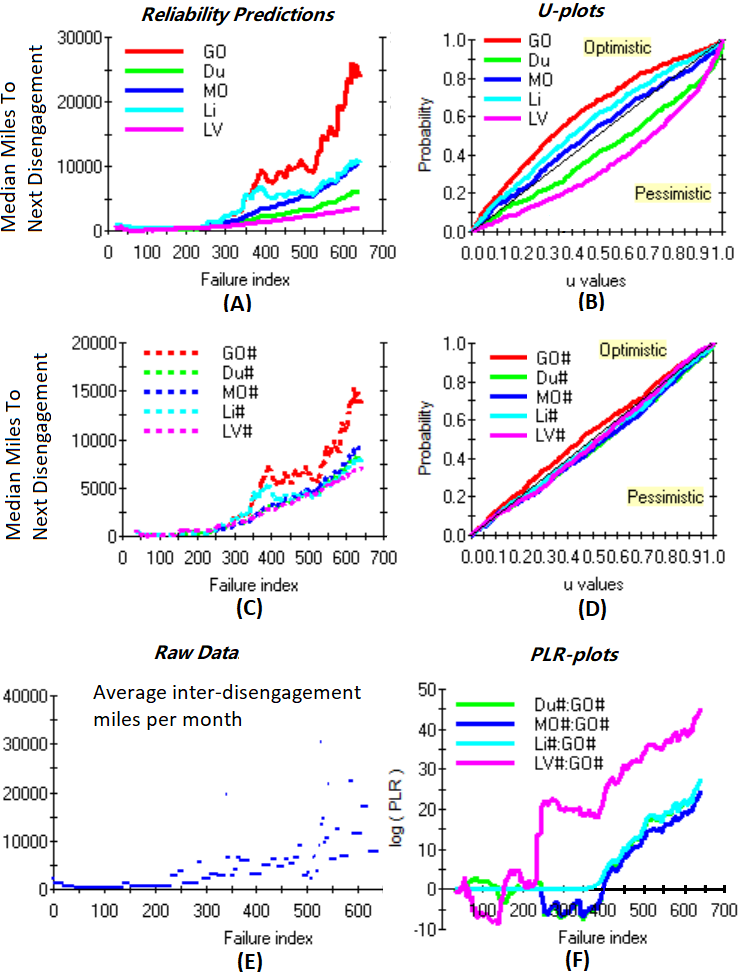}
	\caption{MMTD predictions (A, improved in C), u-plots (B and D), PLR plots for SRGMs (and recalibrated SRGMs), applied to Waymo's 51-month dataset. The SRGMs (plots in A and C) extract predictions about how the trends will continue from the raw data (E); their predictive accuracy can be judged using the other plots. }
	\label{fig_SRGM_for_Waymo}
\end{figure}

In Fig.~\ref{fig_SRGM_for_Waymo}A,C,E,F, the 528 failures in Waymo's disengagement data are indexed in chronological order on the x-axis\footnote{The raw data are  numbers of disengagements, and miles driven, per month; PETERS requires a sequence of inter-failure miles. We preprocessed the raw data by generating random points in a Poisson Process for each month, repeating to check sensitivity of the results to this manipulation.}. 
Fig.~\ref{fig_SRGM_for_Waymo}A shows the successive MMTD predictions (for a better illustration, we show the results of 5 out of the 8 SRGMs implemented in PETERS)\footnote{We chose a set with different enough results to illustrate the method. The abbreviations represent, in order, the SRGMs known as Goel-Okumoto, Duane, Musa-Okumoto, Littlewood, Littlewood-Verrall \cite{littlewood_new_1992}.}. As is common, the SRGMs disagree: GO is more optimistic; LV and Li are more pessimistic.
To check whether they are \textit{objectively} optimistic or pessimistic, we use PETERS' \textit{u-plot} feature. U-plots show how ``unbiased'' a set of predictions is: how close the confidence associated to each prediction is to its actual probability of being correct.
A point on a u-plot, for a value $x$ on the $x$ axis, indicates the fraction of predictions for which the predicted probability of the inter-failure miles that \textit{were observed} was no greater than $x$.
The better calibrated a set of predictions is, the closer the u-plot will be to the diagonal \cite{littlewood_new_1992}.

Fig.~\ref{fig_SRGM_for_Waymo}B shows that most SRGMs proved indeed systematically too optimistic or pessimistic.
The MO predictions seem the best calibrated; however, a good u-plot does not guarantee an SRGM is accurate (or useful) in every way.
Next, to reduce bias, we ``recalibrate'' all models.
Recalibration may improve prediction accuracy (in Fig.~\ref{fig_SRGM_for_Waymo}, the \# suffix identifies a recalibrated model). Fig.~\ref{fig_SRGM_for_Waymo}C shows that recalibration reduced the disagreement between MMTD predictions.
Fig.~\ref{fig_SRGM_for_Waymo}D shows that recalibration drastically reduced bias for most SRGMs 
(MO\# has slightly more pessimistic bias than MO).

To compare these series of predictions by overall accuracy, we use \textit{PLR-plots} (Fig.~\ref{fig_SRGM_for_Waymo}F).
Suppose that two predictors (SRGMs), A and B, give probability density functions $\hat g_j^A(\cdot)$ and $\hat g_j^B(\cdot)$ for the unknown miles to the next failure, given the series of inter-failure miles up to failure $j$.
When a failure does happen, at $m_{j+1}$,
if A is the more accurate predictor, then the ratio $\hat g_{j}^A(m_{j+1})/\hat g_{j}^B(m_{j+1})$ tends to be larger than 1. The PLR of A relative to B (``A:B'' in Fig.~\ref{fig_SRGM_for_Waymo}F)   
is defined as the running product of such ratios, 
$\prod_1^k \left(\hat g_{i-1}^A(m_i)/\hat g_{i-1}^B(m_i)\right)$.
If it consistently increases, then A is generally more accurate than B. 
The PLR-plots in Fig.~\ref{fig_SRGM_for_Waymo}F show that, the four SRGMs that roughly agreed in the MTTD predictions in Fig.\ref{fig_SRGM_for_Waymo}C were, after the 400th failure, generally more accurate (by the same amount: same slope of their PLR-plots) than GO\#, an outlier towards optimism in Fig.\ref{fig_SRGM_for_Waymo}C. For this data set, the best estimate of current MMTD (Fig.~\ref{fig_SRGM_for_Waymo}C) is thus about 7-8000 miles.
%
%
%
%
SRGMs are \textit{not} suitable for deciding whether a safety-critical system satisfies requirements (like those for AVs) of very low rates of serious failures. Even if a SRGM's ``accuracy'' and ``calibration'' properties have proved good, this cannot give high confidence in the one prediction that matters, the one after the \textit{latest} change; that change could have departed from the previous trend -- even radically increasing the failure rate -- but the SRGM would not ``notice'' until the next failure. Yet, SRGMs can be a practical management tool for predicting future inter-event intervals, given large amounts of data, as is the case here for \textit{dpm}. 
By contrast, the CBI developed in Section~\ref{sec_CBI} provides a rigorous approach for safety claims about AVs in scenarios with rare failures.


\section{Related Work}
\label{sec_related_work}

CBI was initially developed for assessing the reliability of conventional safety-critical software in \cite{bishop_toward_2011}. Several extensions, e.g. \cite{strigini_software_2013,zhao_modeling_2017,zhao_conservative_2015,zhao_conservative_2018}, have been developed, considering different prior knowledge and objective functions. CBI has recently been used for estimating catastrophic failure related parameters in the runtime verification of robots \cite{zhao_probabilistic_2019}.

For conventional software, many SRGMs have been developed \cite{min_software_1991}. To the best of our knowledge, the only SRGM developed specifically for ML-based software is \cite{bastani_software_1993}, in which the MO-model was modified to incorporate certain features of AI software.
Differently from \cite{bastani_software_1993}, our approach is to not trust any specific SRGM, but to assess forecast accuracy, improve forecasts, and identify the best SRGMs for the given data.\lorenzo{LS would like something more accurate than ``trust''}


Studies in \cite{banerjee_hands_2018,lv_analysis_2018,favaro_examining_2017,favaro_autonomous_2018,dixit_autonomous_2016} provide descriptive statistics on AV safety and reliability. 
Both \cite{kalra_driving_2016} and \cite{koopman_autonomous_2017} conclude that road testing alone is inadequate evidence of AV safety, and argue the need for alternative methods to supplement public road testing. We agree, and our CBI approach provides a concrete way to incorporate such essential prior knowledge into the assessment.

\xingyu{I added in the last two sentences..}

\section{Conclusions \& Future Work}
\label{sec_conclusions}

The use of machine learning (ML) solutions in safety-critical applications is on the rise. This imposes new challenges on safety and reliability assessment.
For ML systems, the inability to directly verify that a design matches its requirements, by reference to the process of deriving the former from the latter, makes it even harder (compared to conventional software) to estimate the probabilities and consequences of failure  \cite{johnson_increasing_2018}\xingyu{is this sentence too long?}. Thus, we believe, increased reliance on operational testing to study failure probabilities and consequences is inevitable.

In the case of AVs, the problem is also one of demonstrating ``ultra-high reliability'' \cite{littlewood_validation_1993}, for which it is well-known that convincing arguments based on operational testing \textit{alone} are infeasible.
While Bayesian inference supports combining operational testing with other forms of evidence, this latter evidence would need to be such as to support very strong prior beliefs.
Use of safety subsystems -- not relying on the AV's core ML-based systems -- that are verifiable with conventional methods so as to support stronger prior beliefs (than can be had for the ML-based primary system), provides part of the solution. How to support prior beliefs strong enough to give sufficient posterior confidence in the kind of dependability levels now desired for AVs remains an unsolved problem.

Our CBI approach removes the other major difficulty with these problems, that of trusting more detailed prior beliefs than the evidence typically allows one to argue.
One can, thus, take advantage of Bayesian combination of evidence (even given few or no failures) while avoiding possible optimistic bias. This does not solve all of the problems of assessing ``ultra-high dependability'', but it does allow one to trust Bayesian inference; which will deliver enough confidence when requirements are not so extreme (\emph{cf} Fig. \ref{fig_less_stringent_claim}).
For non-ultra-high reliability measures that exhibit growth due to ``learning'' over time, SRGMs, with accuracy validation/recalibration techniques, are useful (at least to derive prior beliefs for inference about reliability of a current version of the AV).

We demonstrate CBI and SRGM methods on one of the most visible examples of an ML-based system with safety-assessment challenges -- autonomous vehicles. To recap, the main contributions of this paper are:

\textit{a) for the assessment of constant, low event rates} -- which is a crucial need for safety claims about AVs -- we propose the ``conservative Bayesian inference'' (CBI) approach.
This approach will be most useful when there are sound bases for prior beliefs, e.g. through safety-oriented architectures in which the ML-based system functions are paired with non-ML safety subsystems, where such safety subsystems are sufficient to avoid accidents and can be rigorously verified. Being a Bayesian approach, CBI allows one to ``give credit'' for this essential evidence.
It can thus contribute to overcoming the challenges of supporting extreme reliability claims; while its conservatism avoids the potential for dangerous errors in the direction of optimism, inherent in common shortcuts for applying Bayes in these cases.


\textit{b) for extrapolating past disengagement trends}, we demonstrate an application of SRGMs to real AV data, with the methods introduced by \cite{brocklehurst_recalibrating_1990,brocklehurst_techniques_1996}. Like previous studies on SRGMs, this example emphasises the importance of continuously evaluating forecasting accuracy, as various applications have shown that no particular SRGM should be expected to always give the ``best'' predictions. Even when an SRGM is shown to outperform others, so far, in a sequence of forecasts, such dominance has been known to change with further observations. We also illustrate how systematic shortcomings in past predictive accuracy can be used to, possibly, improve the performance of these models by using recalibration techniques. This is important with AV reliability data, given AVs' evolving/learning nature and the need to drive under (constantly) changing conditions/environments. The methods for evaluation and recalibration are very general; in principle, they may be applied more widely to point processes. \kizito{shrink/tidy up}


In future work, we plan to explore: 
(a) methods for rigorous claims based on road testing in diverse environments (e.g. cities, traffic regimes; including the case that road testing is ``stratified'' with more testing in those conditions that are expected to be more challenging, while the scenario considered here is of testing that statistically matches expected use);
(b) assessing any alternative models for reliability growth in ML-based systems, in case they prove to deliver more accurate predictions, and studying their possible role in arguments for high reliability;
(c) adapting CBI extensions 
to support sound decisions about the progressive introduction of AVs \cite{strigini_software_2013}.


\lorenzo{this won't fit!!  Just noted here so Kizito can give an opinion, to use in the journal paper...   One of the reviewers asked whether one can assess whether the Bernoulli trial assumption may cause excessive prediction error. This question affects all kinds of inference from test/operation records and has been addressed repeatedly, inlcuing by the current authors \cite{....,....}.  A possible response is that in practice, reasoning about failures per mission may remove parts of the difficulty \cite{...}; or that given sufficiently rare failures, the real failure process and the modelled one will be indistinguishable \cite{Littlewood;  gaudel } but the empirical assessment whether these conditions are satisfied in practice is -- for very reliable systems -- typically lacking and possible infeasible.}

Although we have focused on the ``hot'' area of AVs, our discussion and the novel CBI theorems are more generally applicable. We see them as especially useful now for ML-based systems with critical applications, although not with extreme requirements, since assurance in these systems must rely on combinations of statistical evidence with  other verification methods that are, as yet, not well-established. 

\appendix

\subsection{Statement And Proof of CBI Theorem \ref{thrm_1}}
\label{sec_app_A}
\noindent \textbf{Problem}: 
		Consider the set ${\mathcal D}$ of all probability distributions defined over the unit interval, each distribution representing a potential prior distribution of \emph{pfm} values for an AV. For $0<p_l<\epsilon\leqslant 1$,
		we seek a prior distribution that minimises the posterior confidence in a reliability bound $p\in[p_l, 1]$, given $k$ fatalities have occurred over $n$ miles driven and subject to constraints on some quantiles of the prior distribution. That is, for $\theta\in(0,1]$, we solve 
		\begin{equation*}
		\begin{aligned}
		& \underset{{\mathcal D}}{\text{minimise}}
		& & Pr(X \leqslant p\mid k\&n) \\
		& \text{subject to}
		& & Pr(X \leqslant \epsilon)=\theta,\quad Pr(X\geqslant p_l)=1
		\end{aligned}
		\end{equation*}
	
	\noindent \textbf{Solution}: There is a prior in ${\mathcal D}$ that minimises the posterior confidence: the 2-point distribution $$Pr(X=x)=\theta{\bf 1}_{x=x_1} + (1-\theta){\bf 1}_{x=x_3}$$ where $p_l\leqslant x_1 \leqslant \epsilon < x_3\,$, and the values of $x_1$ and $x_3$ both depend on the model parameters (i.e. $p_l, \epsilon, p$) as well as $k$ and $n$. Using this prior, the minimum posterior confidence is
	\begin{align}
	\frac{x_1^k(1-x_1)^{n-k}\theta}{x_1^k(1-x_1)^{n-k}\theta + x_3^k(1-x_3)^{n-k}(1-\theta)}{\bf 1}_{p>\epsilon}
	\label{eq_res_CBI_post_conf_bound_see_failures_app}
	\end{align}
	where ${\bf 1}_{\tt S}$ is an indicator function -- it is equal to 1 when {\tt S} is true and 0 otherwise.\newline

\begin{proof} The proof is constructive, starting with \emph{any} feasible prior distribution and progressing in 3 stages, each stage producing priors that give progressively worse posterior confidence than in the previous stage. In more detail, assuming $\epsilon \leqslant p$ (the argument for $p<\epsilon$ is analogous):
\begin{enumerate}
    \item First we show that, for any given feasible prior distribution in $\mathcal D$, there is an  equivalent feasible 3-point prior distribution. ``Equivalent'', in that the 3-point distribution has the same value for the posterior confidence in $p$ as the given feasible prior. Consequently, we restrict the optimisation to the set ${\mathcal D}^\ast$ of all such 3-point distributions;
    \item For each prior in ${\mathcal D}^\ast$, there exists a 2-point prior distribution with a smaller posterior confidence in $p$. Consequently, we restrict the optimisation to the set ${\mathcal D}^{\ast\ast}$ of all such 2-point priors;
    \item A monotonicity argument determines a 2-point prior in ${\mathcal D}^{\ast\ast}$ with the smallest posterior confidence in $p$. 
\end{enumerate}

\emph{Stage 1}: Assuming $\epsilon \leqslant p$, note that for any prior distribution $F\in{\mathcal D}$, we may write
	\begin{align}
	\label{eq_appendix_ob}
	Pr(X \leqslant p\mid k\&n)=\frac{T }{T+\int_{p^+}^{1} x^k(1-x)^{n-k} \mathrm dF(x)}
	\end{align}
	where $T\!=\!\int_{p_l}^{\epsilon} x^k(1\!-\!x)^{n-k} \mathrm dF(x)+\!\int_{\epsilon^+}^{p} x^k(1\!-\!x)^{n-k} \mathrm dF(x)$. The \emph{mean-value-theorem for integrals} ensures that three points exist, $x_1\in[p_l,\epsilon]$, $x_2\in(\epsilon,p]$ and $x_3\in (p,1]$, such that \eqref{eq_appendix_ob} becomes (denote $\int_{\epsilon^+}^{p}\mathrm dF(x)=\beta$):
	\begin{align}
	\label{eq_appendix_ob2}
	\frac{x_1^k(1-x_1)^{n-k}\theta+x_2^k(1-x_2)^{n-k}\beta }
	{x_1^k(1\!-\!x_1)^{n\!-\!k}\theta\!+\!x_2^k(1\!-\!x_2)^{n\!-\!k}\beta\!+\!x_3^k(1\!-\!x_3)^{n\!-\!k}(1\!-\!\theta\!-\!\beta)}
	\end{align}

	By establishing \eqref{eq_appendix_ob2} we have established that, for \emph{any} given prior distribution one might start off with, there exists an equivalent 3-point prior distribution. Thus, we restrict the optimisation to ${\mathcal D}^\ast$, the set of all of these equivalent priors. 
	
	\emph{Stage 2:} Next, for each prior in ${\mathcal D}^\ast$, there is a 2-point prior distribution that is guaranteed to give a smaller posterior confidence in $p$. To see this for any given prior in ${\mathcal D}^\ast$ with posterior \eqref{eq_appendix_ob2}, treat all of the other variables as fixed (i.e. the ``$x$''s and $\theta$) and consider which of the allowed values for $\beta$, given these fixed values of the other variables, guarantees a distribution that reduces the posterior confidence. The continuous differentiability of rational functions -- of which \eqref{eq_appendix_ob2} is one -- allows the partial derivative of \eqref{eq_appendix_ob2} w.r.t. $\beta$ to show us the way to do this. The partial derivative of \eqref{eq_appendix_ob2} with respect to $\beta$ is always positive, irrespective of the fixed values the $x_i$s take in their respective ranges. So, to minimise \eqref{eq_appendix_ob2}, we set $\beta=0$. This gives the attainable lower bound \eqref{eq_CBI_post_conf_bound_see_failures_app}, attained by the 2-point prior distribution with probability masses $\theta$ at $x=x_1$, and $1-\theta$ at $x=x_3$. Therefore, we restrict the optimisation to ${\mathcal D}^{\ast\ast}$ -- the set of all such priors.
	\begin{align}
	Pr(\!X \!\leqslant\! p \mid k\&n) &\geqslant  \frac{x_1^k(1-x_1)^{n-k}\theta }{x_1^k(1-x_1)^{n-k}\theta\!+\!x_3^k(1-x_3)^{n-k}(1-\theta)}\nonumber \\
	&= \frac{1}{1 + \left(\frac{x_3^k(1-x_3)^{n-k}}{x_1^k(1-x_1)^{n-k}}\right)\frac{1-\theta}{\theta}}
	\label{eq_CBI_post_conf_bound_see_failures_app}
	\end{align}

\emph{Stage 3:} To minimise \eqref{eq_CBI_post_conf_bound_see_failures_app} further (and, thereby, obtain optimal priors in ${\mathcal D}^{\ast\ast}$), we maximise $x_3^k(1-x_3)^{n-k}$ and minimise $x_1^k(1-x_1)^{n-k}$ over the allowed ranges for $x_1,x_3$. The problem is now reduced to a simple monotonicity analysis given different values of the other model parameters, as follows. Since $x^k(1-x)^{n-k}$ is bell-shaped over $[0,1]$ with a maximum at $x=k/n$, the following defines 2-point priors that solve the optimisation problem (depicted in Fig~\ref{fig_all_priors}):
	
\begin{itemize}
	\item When $0\leqslant k/n \leqslant p_l$:
		\subitem to minimise $x_1^k(1-x_1)^{n-k}$, subject to $x_1\in[p_l,\epsilon]$, we set $x_1=\epsilon$;
		\subitem to maximise $x_3^k(1-x_3)^{n-k}$, subject to $x_3\in (p,1]$, we set $x_3=p$.
		\item When $p_l<k/n\leqslant \epsilon$, and $p_l^k(1-p_l)^{n-k} \geqslant \epsilon^k(1-\epsilon)^{n-k}$:
		\subitem to minimise $x_1^k(1-x_1)^{n-k}$, subject to $x_1\in[p_l,\epsilon]$, we set $x_1=\epsilon$;
		\subitem to maximise $x_3^k(1-x_3)^{n-k}$, subject to $x_3\in (p,1]$, we set $x_3=p$.
		\item When $p_l<k/n\leqslant \epsilon$, and $p_l^k(1-p_l)^{n-k} < \epsilon^k(1-\epsilon)^{n-k}$:
		\subitem to minimise $x_1^k(1-x_1)^{n-k}$, subject to $x_1\in[p_l,\epsilon]$, we set $x_1=p_l$;
		\subitem to maximise $x_3^k(1-x_3)^{n-k}$, subject to $x_3\in (p,1]$, we set $x_3=p$.
		\item When $\epsilon < k/n \leqslant p$:
		\subitem to minimise $x_1^k(1-x_1)^{n-k}$, subject to $x_1\in[p_l,\epsilon]$, we set $x_1=p_l$;
		\subitem to maximise $x_3^k(1-x_3)^{n-k}$, subject to $x_3\in (p,1]$, we set $x_3=p$.
		\item When $p < k/n \leqslant 1$:
		\subitem to minimise $x_1^k(1-x_1)^{n-k}$, subject to $x_1\in[p_l,\epsilon]$, we set $x_1=p_l$;
		\subitem to maximise $x_3^k(1-x_3)^{n-k}$, subject to $x_3\in (p,1]$, we set $x_3=k/n$.
\end{itemize}
	
\begin{figure*}[ht]
		\centering
		\includegraphics[width=0.8\linewidth]{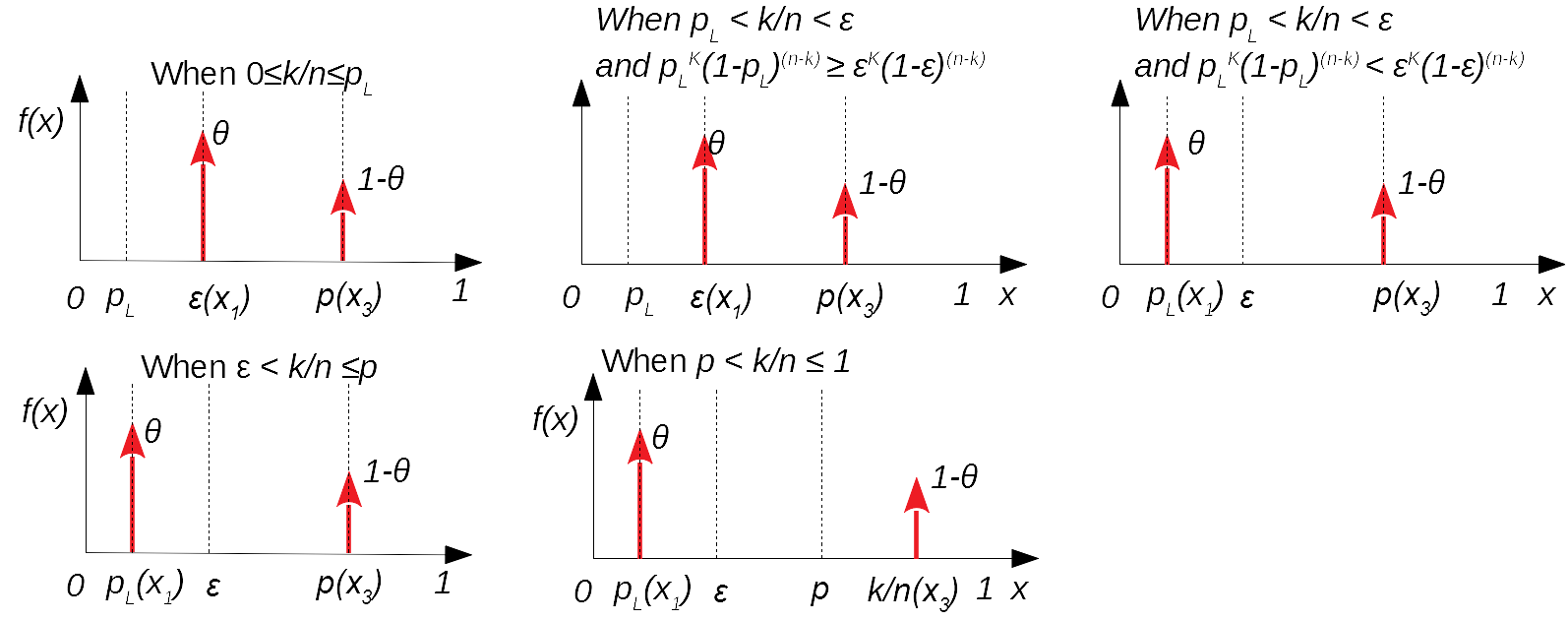}
		\caption{The 5 possible cases of two-point prior distributions that minimise \eqref{eq_appendix_ob}. Notice the important role of where $k/n$ lies.}
		\label{fig_all_priors}
\end{figure*}
\noindent Each prior above has the form \eqref{eq_res_CBI_post_conf_bound_see_failures_app} for $Pr(X \leqslant p\mid k\&n)$.

All of the foregoing proves Theorem \ref{thrm_1} for $\epsilon\leqslant p$. Begin the optimisation again, but now assuming $p<\epsilon$. For any feasible prior $F\in{\mathcal D}$, the objective function $Pr(X \leqslant p\mid k\&n)$ can be written as
\begin{align}
\label{eq_appendix_ob3}
\frac{L}{L+\int_{p^+}^{\epsilon} x^k(1-x)^{n-k} \mathrm dF(x)+\int_{\epsilon^+}^{1} x^k(1-x)^{n-k} \mathrm dF(x)}
\end{align}

where $L=\int_{p_l}^{p} x^k(1-x)^{n-k} \mathrm dF(x)$. As before, the \emph{mean-value-theorem} ensures the existence of three points $x_1,x_2,x_3$ in the ranges: $x_1\in [p_l,p], x_2\in (p,\epsilon],x_3\in (\epsilon,1]$ such that \eqref{eq_appendix_ob3} becomes (denote $\int_{p_l}^{p}\mathrm dF(x)=\gamma$, where $0 \leqslant \gamma \leqslant \theta$):
\begin{align}
\label{eq_appendix_ob4}
\frac{L'}
{L'+x_2^k(1-x_2)^{n-k}(\theta-\gamma)+x_3^k(1-x_3)^{n-k}(1-\theta)}
\end{align}
where $L'=x_1^k(1-x_1)^{n-k}\gamma$.

The derivative of \eqref{eq_appendix_ob4} with respect to $\gamma$ is always positive, irrespective of the fixed values the $x_i$s can take in their allowed ranges. So, to minimise \eqref{eq_appendix_ob4}, we simply set $\gamma=0$. Thus, \eqref{eq_appendix_ob4} has an attainable bound of 0 when $p<\epsilon$, and the corresponding prior distribution that attains this is still a 2-point one with probability masses at $x=x_2$ and $x=x_3$, regardless of what fixed values $x_2$ and $x_3$ take in their allowed ranges.
\end{proof}

\subsection{Formal Analysis for Q3 in Sec.~\ref{sec_num_examples_CBI} }
\label{sec_app_B}

We seek to understand what happens when $n_1$ fatality-free driven miles support a \emph{pfm} claim $p$ with confidence $c$. And, upon seeing a fatality after $n_1$ miles, understanding how many more fatality-free miles $n_2$ are needed to maintain support for the claim. So, what follows is an analysis of the asymptotic ``large $n$'' behaviour implied by the worst-case posterior confidence \eqref{eq_CBI_post_conf_bound_see_failures} in Theorem \ref{thrm_1}. Assume $c$ and $\theta$ are given in the practical case when $c\geqslant \theta$.

Let $n^\ast$ denote the number of miles that satisfies $\epsilon(1-\epsilon)^{n^\ast-1} = p_l(1-p_l)^{n^\ast-1}$. So, from appendix \ref{sec_app_A} above, for $n< n^\ast$ we have $x_1=p_l$, and for $n\geqslant n^\ast$ we have $x_1=\epsilon$. Note that $n^\ast$ is independent of $c$ and $\theta$, so this number of miles will be the same no matter what levels of confidence one is either interested in, or has prior to road testing.

Now, using \eqref{eq_CBI_post_conf_bound_see_failures}, we may write the number of miles driven as a function of the remaining problem parameters. That is, for $\epsilon<p\leqslant 1$,
\begin{equation}
n(c, p, \theta, x_1, k) := k + \left(\frac{k\log(x_1/p) + \log(\frac{\theta(1-c)}{c(1-\theta)})}{\log(\frac{1-p}{1-x_1})}\right)
\label{eq_n_formula}
\end{equation}
where we have assumed that the values of $n$ ensure $k/n\leqslant p$ holds. In particular, for $k=1$, let $p^\ast$ uniquely satisfy
\begin{equation}
n^\ast = 1 + \left(\frac{\log(x_1/p^\ast) + \log(\frac{\theta(1-c)}{c(1-\theta)})}{\log(\frac{1-p^\ast}{1-x_1})}\right)
\label{eq_pstar}
\end{equation}
where $x_1=p_l, \epsilon$ both result in the same $n^*$ value, by the definition of $n^\ast$. So, for $p>p^\ast$, we must have $x_1=p_l$. And, for $\epsilon < p\leqslant p^\ast$, we have $x_1=\epsilon$.

If, for otherwise fixed parameter values, we denote $\tilde{n}$ the number of miles according to \eqref{eq_n_formula} when $k=1$, and $n_1$ the number of miles when $k=0$, then the number of additional miles $n_2$ needed upon seeing a fatality immediately after $n_1$ miles is $n_2 := \tilde{n} - n_1$.

Suppose then, that $p>p^\ast$ and let $p$ tend to $p^\ast$ from above. The following limits follow from the continuity of $n$ in \eqref{eq_n_formula}:
\begin{enumerate}
	\item \emph{If a fatality is observed (so $k=1$)} then, as $p$ tends to $p^\ast$ from above, we have $x_1 = p_l$, and the number of miles that are needed to be driven to support a claim in $p$ -- with confidence $c$ using prior confidence $\theta$ in the engineering goal $\epsilon$ being met -- is 
	\begin{align*}
	\lim\limits_{p\downarrow p^\ast}\tilde{n}& = \lim\limits_{p\downarrow p^\ast} n(c, p, \theta, p_l, 1)  \\
	&= n(c, \lim\limits_{p\downarrow p^\ast} p, \theta, p_l, 1) = n(c, p^\ast, \theta, p_l, 1) = n^\ast
	\end{align*}
	
	\item \emph{If no fatalities are observed (so $k=0$)} then, as $p$ tends to $p^\ast$ from above, the number of fatality-free miles that are needed to be driven to support a claim in $p$ -- with confidence $c$ using prior confidence $\theta$ in the engineering goal $\epsilon$ being met -- is 
	\begin{align*}
	\lim\limits_{p\downarrow p^\ast}n_1& = \lim\limits_{p\downarrow p^\ast} n(c, p, \theta, \epsilon, 0) =n(c, \lim\limits_{p\downarrow p^\ast} p, \theta, \epsilon, 0) \\
	& =n(c, p^\ast, \theta, \epsilon, 0) = \frac{ \log(\frac{\theta(1-c)}{c(1-\theta)})}{\log(\frac{1-p^\ast}{1-\epsilon})}
	\end{align*}
	\noindent Recall, from appendix \ref{sec_app_A}, that $x_1 = \epsilon$ must hold here for all $p$ when $k=0$.
	
	\item so, using these last two results, the number of extra miles needed is 
	\begin{equation}
	\lim\limits_{p\downarrow p^\ast}n_2 = n^\ast - \frac{ \log(\frac{\theta(1-c)}{c(1-\theta)})}{\log(\frac{1-p^\ast}{1-\epsilon})}
	\label{eq_n2fromabove}
	\end{equation}
\end{enumerate}

Alternatively, suppose $p<p^\ast$ and let $p$ tend to $\epsilon$ from above. The following limits also follow from \eqref{eq_n_formula}:
\begin{enumerate}
	\item \emph{If a fatality is observed (so $k=1$)}, then as $p$ tends to $\epsilon$ from above, we have $x_1 = \epsilon$, and the number of miles that are needed to be driven to support a claim in $p$ -- with confidence $c$ using prior confidence $\theta$ in the engineering goal $\epsilon$ being met -- is 
	\begin{equation*}
	\lim\limits_{p\downarrow \epsilon}\tilde{n} = \lim\limits_{p\downarrow \epsilon} n(c, p, \theta, \epsilon, 1) =  n(c, \lim\limits_{p\downarrow \epsilon} p, \theta, \epsilon, 1) = \infty
	\end{equation*}
	
	\item \emph{If no fatalities are observed (so $k=0$)} then, as $p$ tends to $\epsilon$ from above, the number of fatality-free miles that are needed to be driven to support a claim in $p$ -- with confidence $c$ using prior confidence $\theta$ in the engineering goal $\epsilon$ being met -- is 
	\begin{equation*}
	\lim\limits_{p\downarrow \epsilon}n_1 = \lim\limits_{p\downarrow \epsilon} n(c, p, \theta, \epsilon, 0) =  n(c, \lim\limits_{p\downarrow \epsilon} p, \theta, \epsilon, 0) = \infty
	\end{equation*}
	
	\item the last two results show that both $\tilde{n}$ and $n_1$ grow without bound, however the number of extra miles needed is bounded above, since (by \emph{L'Hospital's rule})
	\begin{align}
	\lim\limits_{p\downarrow \epsilon}n_2 &= \lim\limits_{p\downarrow \epsilon}(\tilde{n} -n_1) 	\nonumber\\
	&=\lim\limits_{p\downarrow \epsilon} ( n(c, p, \theta, \epsilon, 1) - n(c, p, \theta, \epsilon, 0) )  \nonumber\\
	& =1 + \lim\limits_{p\downarrow \epsilon}\left(\frac{\log(\epsilon/p)}{\log(\frac{1-p}{1-\epsilon})}\right) \nonumber \\
	&= 1 + \lim\limits_{p\downarrow \epsilon}\frac{(1/p)}{1/(1-p)} = 1 + \frac{1-\epsilon}{\epsilon} = 1/\epsilon
	\label{eq_n2pastpstar}
	\end{align}
	\noindent Note that, like $n^\ast$, this limit is independent of $c$ and $\theta$.
\end{enumerate}

\bibliographystyle{IEEEtran}
\bibliography{ref}

\begin{thebibliography}{10}
\providecommand{\url}[1]{#1}
\csname url@samestyle\endcsname
\providecommand{\newblock}{\relax}
\providecommand{\bibinfo}[2]{#2}
\providecommand{\BIBentrySTDinterwordspacing}{\spaceskip=0pt\relax}
\providecommand{\BIBentryALTinterwordstretchfactor}{4}
\providecommand{\BIBentryALTinterwordspacing}{\spaceskip=\fontdimen2\font plus
\BIBentryALTinterwordstretchfactor\fontdimen3\font minus
  \fontdimen4\font\relax}
\providecommand{\BIBforeignlanguage}[2]{{%
\expandafter\ifx\csname l@#1\endcsname\relax
\typeout{** WARNING: IEEEtran.bst: No hyphenation pattern has been}%
\typeout{** loaded for the language `#1'. Using the pattern for}%
\typeout{** the default language instead.}%
\else
\language=\csname l@#1\endcsname
\fi
#2}}
\providecommand{\BIBdecl}{\relax}
\BIBdecl

\bibitem{anderson_autonomous_2016}
J.~M. Anderson, K.~Nidhi, K.~D. Stanley, P.~Sorensen, C.~Samaras, and O.~A.
  Oluwatola, ``Autonomous vehicle technology: {A} guide for policymakers,''
  Rand Corporation, Tech. Rep. RR-443-2-RC, 2016.

\bibitem{paden_survey_2016}
B.~Paden, M.~Čáp, S.~Z. Yong, D.~Yershov, and E.~Frazzoli, ``A survey of
  motion planning and control techniques for self-driving urban vehicles,''
  \emph{IEEE Tran. on Intelligent Vehicles}, vol.~1, no.~1, pp. 33--55, 2016.

\bibitem{fagnant_preparing_2015}
D.~J. Fagnant and K.~Kockelman, ``Preparing a nation for autonomous vehicles:
  {Opportunities}, barriers and policy recommendations,'' \emph{Transp.
  Research Part A: Policy and Practice}, vol.~77, pp. 167--181, 2015.

\bibitem{koopman_autonomous_2017}
P.~Koopman and M.~Wagner, ``Autonomous vehicle safety: {An} interdisciplinary
  challenge,'' \emph{IEEE Intelligent Transportation Systems Magazine}, vol.~9,
  no.~1, pp. 90--96, 2017.

\bibitem{bonnefon_social_2016}
J.-F. Bonnefon, A.~Shariff, and I.~Rahwan, ``The social dilemma of autonomous
  vehicles,'' \emph{Science}, vol. 352, no. 6293, pp. 1573--1576, 2016.

\bibitem{schwarting_planning_2018}
W.~Schwarting, J.~Alonso-Mora, and D.~Rus, ``Planning and decision-making for
  autonomous vehicles,'' \emph{Annual Review of Control, Robotics, and
  Autonomous Systems}, vol.~1, no.~1, pp. 187--210, 2018.

\bibitem{urmson_hands_2016}
C.~Urmson, ``Hands off: {The} future of self-driving cars,'' Committee on
  Commerce, Science and Transportation, Washington, D.C., USA, Testimony, 2016.

\bibitem{banerjee_hands_2018}
S.~S. Banerjee, S.~Jha, J.~Cyriac, Z.~T. Kalbarczyk, and R.~K. Iyer, ``Hands
  off the wheel in autonomous vehicles?: {A} systems perspective on over a
  million miles of field data,'' in \emph{48th {IEEE}/{IFIP} {Int.} {Conf.} on
  {Dependable} {Systems} and {Networks}}, 2018, pp. 586--597.

\bibitem{kalra_driving_2016}
N.~Kalra and S.~Paddock, ``Driving to safety: {How} many miles of driving would
  it take to demonstrate autonomous vehicle reliability?'' \emph{Transp.
  Research Part A: Policy and Practice}, vol.~94, pp. 182--193, 2016.

\bibitem{favaro_autonomous_2018}
F.~Favarò, S.~Eurich, and N.~Nader, ``Autonomous vehicles' disengagements:
  {Trends}, triggers, and regulatory limitations,'' \emph{Accident Analysis \&
  Prevention}, vol. 110, pp. 136 -- 148, 2018.

\bibitem{dixit_autonomous_2016}
V.~V. Dixit, S.~Chand, and D.~J. Nair, ``Autonomous vehicles: {Disengagements},
  accidents and reaction times,'' \emph{PLOS ONE}, vol.~11, no.~12, pp. 1--14,
  2016.

\bibitem{lv_analysis_2018}
C.~Lv, D.~Cao, Y.~Zhao, D.~J. Auger, M.~Sullman, H.~Wang, L.~M. Dutka,
  L.~Skrypchuk, and A.~Mouzakitis, ``Analysis of autopilot disengagements
  occurring during autonomous vehicle testing,'' \emph{IEEE/CAA Journal of
  Automatica Sinica}, vol.~5, no.~1, pp. 58--68, Jan. 2018.

\bibitem{littlewood_validation_1993}
B.~Littlewood and L.~Strigini, ``Validation of ultra-high dependability for
  software-based systems,'' \emph{Comm. of the ACM}, vol.~36, pp. 69--80, 1993.

\bibitem{butler_infeasibility_1993}
R.~W. Butler and G.~B. Finelli, ``The infeasibility of quantifying the
  reliability of life-critical real-time software,'' \emph{IEEE Transactions on
  Software Engineering}, vol.~19, no.~1, pp. 3--12, Jan. 1993.

\bibitem{littlewood_reasoning_2012}
B.~Littlewood and J.~Rushby, ``Reasoning about the reliability of diverse
  two-channel systems in which one channel is `possibly perfect','' \emph{IEEE
  Tran. on Software Engineering}, vol.~38, no.~5, pp. 1178--1194, 2012.

\bibitem{waymo_waymo_2018}
\BIBentryALTinterwordspacing
{Waymo}, ``Waymo safety report: {On} the road to fully self-driving,'' Tech.
  Rep., 2018. [Online]. Available:
  \url{https://storage.googleapis.com/sdc-prod/v1/safety-report/Safety Report
  2018.pdf}
\BIBentrySTDinterwordspacing

\bibitem{amnon_shashua_plan_2017}
\BIBentryALTinterwordspacing
A.~Shashua and S.~Shalev-Shwartz, ``A plan to develop safe autonomous vehicles.
  {And} prove it,'' \emph{Intel Newsroom}, p.~8, 2017. [Online]. Available:
  \url{https://newsroom.intel.com/newsroom/wp-content/uploads/sites/11/2017/10/autonomous-vehicle-safety-strategy.pdf}
\BIBentrySTDinterwordspacing

\bibitem{tian_deeptest_2018}
Y.~Tian, K.~Pei, S.~Jana, and B.~Ray, ``{DeepTest}: {Automated} testing of
  deep-neural-network-driven autonomous cars,'' in \emph{the 40th {Int.}
  {Conf.} on {Software} {Engineering}}, New York, NY, USA, 2018, pp. 303--314.

\bibitem{kamali_formal_2017}
M.~Kamali, L.~A. Dennis, O.~McAree, M.~Fisher, and S.~M. Veres, ``Formal
  verification of autonomous vehicle platooning,'' \emph{Science of Computer
  Programming}, vol. 148, pp. 88 -- 106, 2017.

\bibitem{fisher_verifiable_2018}
M.~Fisher, E.~Collins, L.~Dennis, M.~Luckcuck, M.~Webster, M.~Jump, V.~Page,
  C.~Patchett, F.~Dinmohammadi, D.~Flynn, V.~Robu, and X.~Zhao, ``Verifiable
  self-certifying autonomous systems,'' in \emph{{IEEE} {Int.} {Symp.} on
  {Software} {Reliability} {Engineering} {Workshops}}, 2018, pp. 341--348.

\bibitem{koopman_safety_2019}
P.~Koopman and B.~Osyk, ``Safety argument considerations for public road
  testing of autonomous vehicles,'' in \emph{{WCX} {SAE} {World} {Congress}
  {Experience}}.\hskip 1em plus 0.5em minus 0.4em\relax SAE International, Apr.
  2019.

\bibitem{PopovStrigini2010ISSRE}
P.~Popov and L.~Strigini, ``Assessing asymmetric fault-tolerant software,'' in
  \emph{the 21st Int. Symp. on Software Reliability Engineering}.\hskip 1em
  plus 0.5em minus 0.4em\relax San Jose, CA, USA: IEEE Computer Society Press,
  2010, pp. 41--50.

\bibitem{bishop_toward_2011}
P.~Bishop, R.~Bloomfield, B.~Littlewood, A.~Povyakalo, and D.~Wright, ``Toward
  a formalism for conservative claims about the dependability of software-based
  systems,'' \emph{IEEE Transactions on Software Engineering}, vol.~37, no.~5,
  pp. 708--717, 2011.

\bibitem{strigini_software_2013}
L.~Strigini and A.~Povyakalo, ``Software fault-freeness and reliability
  predictions,'' in \emph{Computer {Safety}, {Reliability}, and {Security}},
  ser. LNCS, vol. 8153.\hskip 1em plus 0.5em minus 0.4em\relax Springer Berlin
  Heidelberg, 2013, pp. 106--117.

\bibitem{zhao_modeling_2017}
X.~Zhao, B.~Littlewood, A.~Povyakalo, L.~Strigini, and D.~Wright, ``Modeling
  the probability of failure on demand (pfd) of a 1-out-of-2 system in which
  one channel is ``quasi-perfect'','' \emph{Reliability Engineering \& System
  Safety}, vol. 158, pp. 230--245, 2017.

\bibitem{zhao_conservative_2015}
X.~Zhao, B.~Littlewood, A.~Povyakalo, and D.~Wright, ``Conservative claims
  about the probability of perfection of software-based systems,'' in
  \emph{26th {Int.} {Symp.} on {Software} {Reliability} {Eng.}}\hskip 1em plus
  0.5em minus 0.4em\relax IEEE, 2015, pp. 130--140.

\bibitem{Miller1986EOS}
D.~R. Miller, ``Exponential order statistic models of software reliability
  growth,'' \emph{IEEE Tran. on Software Eng.}, vol.~12, no.~01, pp. 12--24,
  1986.

\bibitem{brocklehurst_techniques_1996}
S.~Brocklehurst and B.~Littlewood, ``Techniques for prediction analysis and
  recalibration,'' in \emph{Handbook of {Software} {Reliability} {Eng.}},
  M.~Lyu, Ed.\hskip 1em plus 0.5em minus 0.4em\relax McGraw-Hill \& IEEE
  Computer Society Press, 1996, pp. 119--166.

\bibitem{iec_61508_2010}
IEC, \emph{{IEC61508}, {Functional} {Safety} of {Electrical}/
  {Electronic}/{Programmable} {Electronic} {Safety} {Related} {Systems}}, 2010.

\bibitem{en50129_railway_2003}
CENELEC, \emph{EN50129, {Railway} {Applications}-{Communication}, {Signalling}
  and {processing} {Systems}-{Safety} {Related} {Electronic} {Systems} for
  {Signalling}}, 2003.

\bibitem{atwood2003handbook}
C.~Atwood, J.~LaChance, H.~Martz, D.~Anderson, M.~Englehardt, D.~Whitehead, and
  T.~Wheeler, ``Handbook of parameter estimation for probabilistic risk
  assessment,'' U.S. Nuclear Regulatory Commission, Washington, DC, Report
  NUREG/CR-6823, 2003.

\bibitem{strigini_guidelines_1997}
L.~Strigini and B.~Littlewood, ``Guidelines for statistical testing,'' City
  University London, Project {Report} PASCON/WO6-CCN2/TN12, 1997.

\bibitem{walter_bayesian_2017}
G.~Walter, L.~J.~M. Aslett, and F.~P.~A. Coolen, ``Bayesian nonparametric
  system reliability using sets of priors,'' \emph{International Journal of
  Approximate Reasoning}, vol.~80, pp. 67--88, 2017.

\bibitem{bishop_deriving_2017}
P.~Bishop and A.~Povyakalo, ``Deriving a frequentist conservative confidence
  bound for probability of failure per demand for systems with different
  operational and test profiles,'' \emph{Reliability Engineering \& System
  Safety}, vol. 158, pp. 246--253, 2017.

\bibitem{utkin_imprecise_2018}
L.~V. Utkin and F.~P.~A. Coolen, ``Imprecise probabilistic inference for
  software run reliability growth models.'' \emph{Journal of Uncertain
  Systems.}, vol.~12, no.~4, pp. 292--308, 2018.

\bibitem{favaro_examining_2017}
F.~M. Favarò, N.~Nader, S.~O. Eurich, M.~Tripp, and N.~Varadaraju, ``Examining
  accident reports involving autonomous vehicles in {California},'' \emph{PLOS
  ONE}, vol.~12, no.~9, pp. 1--20, 2017.

\bibitem{cinlarStocProcBook}
E.~Cinlar, \emph{Introduction to Stochastic Processes}, ser. Dover Books on
  Mathematics Series.\hskip 1em plus 0.5em minus 0.4em\relax Dover
  Publications, Incorporated, 2013.

\bibitem{zhao_conservative_2018}
X.~Zhao, B.~Littlewood, A.~Povyakalo, L.~Strigini, and D.~Wright,
  ``Conservative claims for the probability of perfection of a software-based
  system using operational experience of previous similar systems,''
  \emph{Reliability Engineering \& System Safety}, vol. 175, pp. 265 -- 282,
  2018.

\bibitem{liu_how_2019}
P.~Liu, R.~Yang, and Z.~Xu, ``How safe is safe enough for self-driving
  vehicles?'' \emph{Risk Analysis}, vol.~39, no.~2, pp. 315--325, 2019.

\bibitem{awad_moral_2018}
E.~Awad, S.~Dsouza, R.~Kim, J.~Schulz, J.~Henrich, A.~Shariff, J.-F. Bonnefon,
  and I.~Rahwan, ``The {Moral} {Machine} experiment,'' \emph{Nature}, vol. 563,
  no. 7729, pp. 59--64, 2018.

\bibitem{berger_could_2003}
J.~O. Berger, ``Could {Fisher}, {Jeffreys} and {Neyman} have agreed on
  testing?'' \emph{Statistical Science}, vol.~18, no.~1, pp. 1--32, 2003.

\bibitem{abdel-ghaly_evaluation_1986}
A.~A. Abdel-Ghaly, P.~Y. Chan, and B.~Littlewood, ``Evaluation of competing
  software reliability predictions,'' \emph{IEEE Transactions on Software
  Engineering}, vol. SE-12, no.~9, pp. 950--967, 1986.

\bibitem{littlewood_new_1992}
S.~Brocklehurst and B.~Littlewood, ``New ways to get accurate reliability
  measures,'' \emph{IEEE Software}, vol.~9, pp. 34--42, 1992.

\bibitem{zhao_probabilistic_2019}
X.~Zhao, V.~Robu, D.~Flynn, F.~Dinmohammadi, M.~Fisher, and M.~Webster,
  ``Probabilistic model checking of robots deployed in extreme environments,''
  in \emph{Proc. of the 33rd {AAAI} {Conference} on {Artificial}
  {Intelligence}}, vol.~33, Honolulu, Hawaii, USA, 2019, pp. 8076--8084.

\bibitem{min_software_1991}
M.~Xie, \emph{Software reliability modelling}.\hskip 1em plus 0.5em minus
  0.4em\relax World Scientific, 1991, vol.~1.

\bibitem{bastani_software_1993}
F.~B. Bastani, I.-R. Chen, and T.-W. Tsao, ``A software reliability model for
  artificial intelligence programs,'' \emph{Int. Journal of Software
  Engineering and Knowledge Engineering}, vol.~3, no.~01, pp. 99--114, 1993.

\bibitem{johnson_increasing_2018}
{Johnson, C. W.}, ``The increasing risks of risk assessment: {On} the rise of
  artificial intelligence and non-determinism in safety-critical systems,'' in
  \emph{the 26th {Safety}-{Critical} {Systems} {Symposium}}.\hskip 1em plus
  0.5em minus 0.4em\relax York, UK.: Safety-Critical Systems Club, 2018, p.~15.

\bibitem{brocklehurst_recalibrating_1990}
S.~Brocklehurst, P.~Y. Chan, B.~Littlewood, and J.~Snell, ``Recalibrating
  software reliability models,'' \emph{IEEE Transactions on Software
  Engineering}, vol.~16, no.~4, pp. 458--470, Apr. 1990.

\end{thebibliography}

\end{document}